\documentclass[conference]{IEEEtran}

\usepackage{cite}
\usepackage{amsmath,amssymb,amsfonts}
\usepackage{graphicx}
\usepackage{textcomp}
\usepackage{xcolor}
\usepackage{booktabs}
\usepackage[hidelinks]{hyperref}

\usepackage{geometry}

\def\BibTeX{{\rm B\kern-.05em{\sc i\kern-.025em b}\kern-.08em
    T\kern-.1667em\lower.7ex\hbox{E}\kern-.125emX}}

\begin{document}

\title{Reference-Free Omnidirectional Stereo Matching via Multi-View Consistency Maximization}

\author{
\IEEEauthorblockN{Lehuai Xu\textsuperscript{1,3},
Weiming Zhang\textsuperscript{2},
Yang Li\textsuperscript{1},
Sidan Du\textsuperscript{1},
Lin Wang\textsuperscript{3,*}}
\IEEEauthorblockA{\textsuperscript{1}Nanjing University, Nanjing, China}
\IEEEauthorblockA{\textsuperscript{2}HKUST(GZ), Guangzhou, China}
\IEEEauthorblockA{\textsuperscript{3}Nanyang Technological University, Singapore}
\IEEEauthorblockA{\textsuperscript{*}Corresponding author: linwang@ntu,edu.sg}
}

\maketitle

\begin{abstract}
Reliable omnidirectional depth estimation from multi-fisheye stereo matching is pivotal to many applications, such as embodied robotics.
Existing approaches either rely on spherical sweeping with heuristic fusion strategies to build the cost columns or perform reference-centric stereo matching based on rectified views.
However, these methods fail to explicitly exploit geometric relationships between multiple views, rendering them less capable of capturing the global dependencies, visibility, or scale changes.
In this paper, we shift to a new perspective and propose a novel reference-free framework, dubbed FreeOmniMVS, via multi-view consistency maximization.
The highlight of FreeOmniMVS is that it can aggregate pair-wise correlations into a robust, visibility-aware, and global consensus.
As such, it is tolerant to occlusions, partial overlaps, and varying baselines.
Specifically, to achieve global coherence, we introduce a novel View-pair Correlation Transformer (VCT) that explicitly models pairwise correlation volumes across all camera view pairs, allowing us to drop unreliable pairs caused by occlusion or out-of-focus observations.
To realize scalable and visibility-aware consensus, we propose a lightweight attention mechanism that adaptively fuses the correlation vectors, eliminating the need for a designated reference view and allowing all cameras to contribute equally to the stereo matching process.
Extensive experiments on diverse benchmark datasets demonstrate the superiority of our method for globally consistent, visibility-aware, and scale-aware omnidirectional depth estimation.
Code will be publicly available.
\end{abstract}


\section{Introduction}
\label{sec:intro}

Multi-camera fisheye systems have become a key sensing modality for applications demanding full surround-view 3D perception, ranging from autonomous driving \cite{hane2014real,cui2019realtime,yogamani2019woodscape} and mobile robot navigation \cite{shah1994depth,kita2013obstacle,li2024hexamode} to unmanned aerial vehicle spatial awareness \cite{gao2017dual,pulling2024geometry}. Central to all these scenarios is omnidirectional multi-view stereo (MVS), which reconstructs a coherent and globally consistent $360^\circ$ depth map by matching features across multiple overlapping fisheye views and recovering metrically accurate absolute depth through geometric triangulation \cite{won2019omnimvs,won2020omnimvs,meuleman2021realtime,xie2023omnividar}. The fundamental challenge of this problem lies not only in estimating accurate depth for each individual view, but also in maintaining consistency across all cameras observing overlapping regions. Even slight inter-view inconsistencies can introduce discontinuities on the reconstructed sphere, undermining the reliability of the resulting $360^\circ$ geometry.

\begin{figure}[t]
  \centering
  \includegraphics[width=0.9\linewidth]{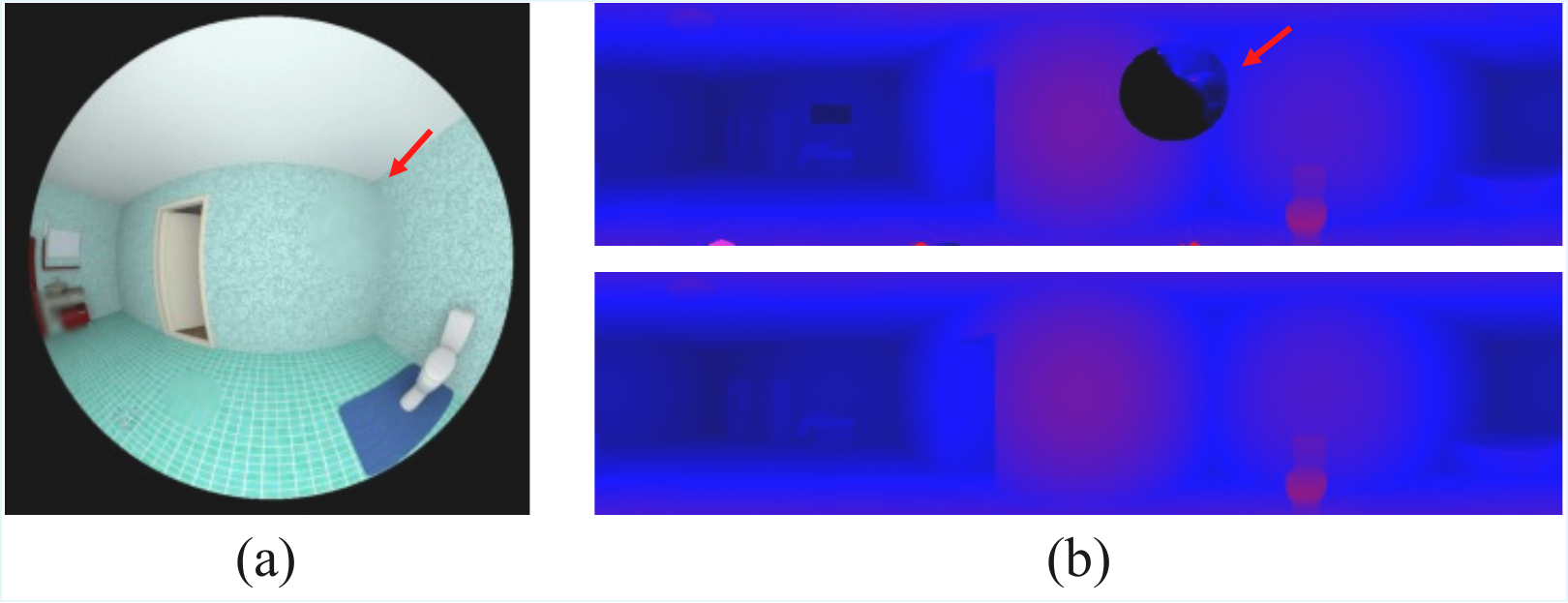}
  \includegraphics[width=0.9\linewidth]{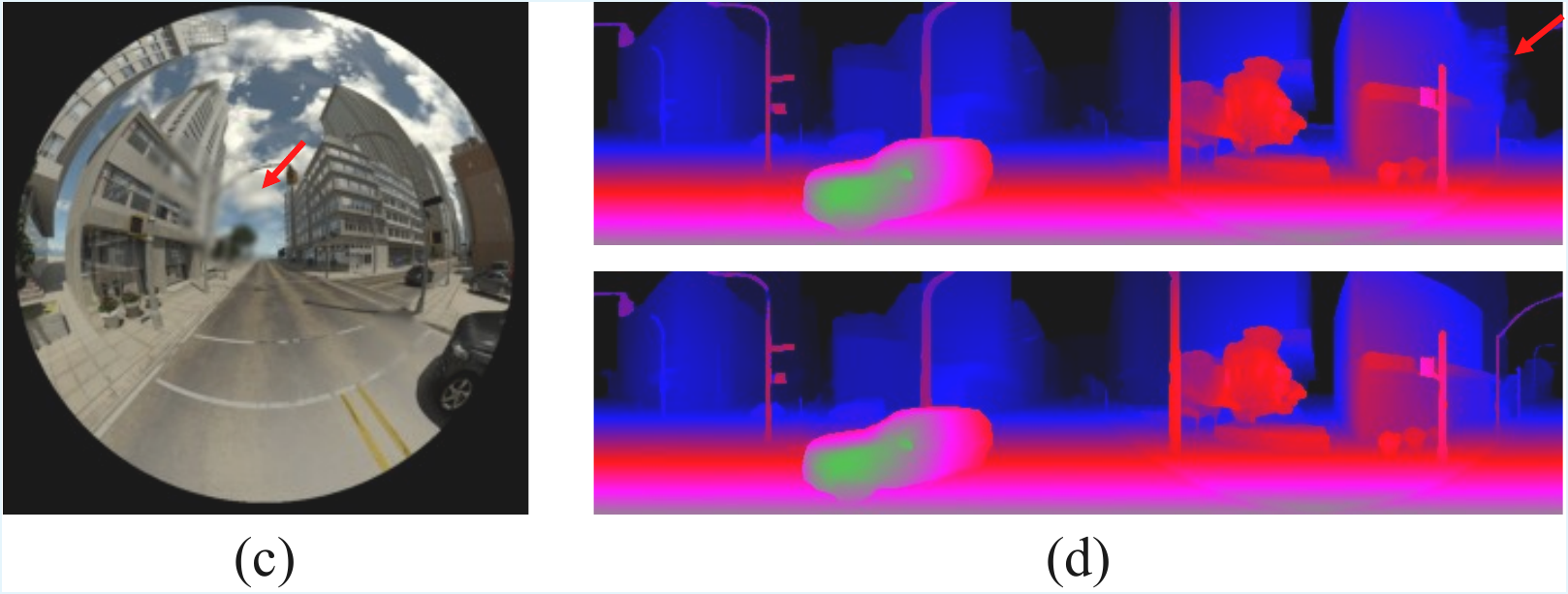}
  \vspace{-10pt}
  \caption{\textbf{(a,c)} Fisheye inputs with randomly injected local blur and noise, simulating out-of-focus or corrupted observations. \textbf{(b,d)} Depth predictions from our backbone (top) versus FreeOmniMVS (bottom). Our method suppresses artifacts around degraded regions and preserves thin structures and cross-camera consistency across the full $360^\circ$ field of view.}
  \label{fig:short}
  \vspace{-12pt}
\end{figure}

Existing omni-depth pipelines mainly pursue two lines. One relies on \emph{spherical sweeping} accompanied by heuristic fusion to build cost columns on concentric spheres \cite{won2019sweepnet,won2020end,komatsu2020360,chen2023s,jiang2024romnistereo}. This path affords geometric interpretability, yet its computation and memory grow rapidly with the number of depth hypotheses and cameras, and the final fusion is typically heuristic. The other performs \emph{stereo matching on rectified views} in a \emph{reference-centric} manner \cite{li2022mode,xie2023omnividar,deng2025omnistereo}, which is efficient and easier to learn but processes pairs independently and privileges a designated reference. Despite their differences, both families fail to explicitly exploit the geometric relationships among multiple views: global dependencies remain implicit, visibility is not handled explicitly, and scale changes or large baselines often destabilize fusion.

We therefore shift to a new perspective and advocate a \emph{reference-free} formulation that seeks \textbf{multi-view consistency maximization}. Concretely, rather than comparing each source to a chosen reference, we compute correlation evidence across all camera pairs and aggregate it into a single, robust consensus per depth and location. Such an explicit, visibility-aware consensus enables the estimator to tolerate occlusions, partial overlaps, and varying baselines while promoting global coherence on the sphere.


To this end, we propose \textbf{FreeOmniMVS}, a novel approach based on a \textbf{View-pair Correlation Transformer} that builds pairwise correlation volumes for all camera combinations and uses an attention mechanism to adaptively fuse the resulting correlation vectors. During training we inject Gumbel perturbations to approximate Top-$k$ selection; at inference we adopt a hard Top-$k$ mask. This allows the model to drop unreliable pairs arising from occlusion or out-of-FoV regions, thereby forming a visibility-aware consensus without relying on any designated reference view. All fisheye cameras contribute equally to the depth estimation process, and the aggregation cost scales controllably with the number of view pairs.

Building on this idea, we adopt the three-stage backbone of RomniStereo \cite{jiang2024romnistereo} and plug our consensus module into its middle stage. Stage 1 follows OmniMVS-style spherical sweeping to extract unary fisheye features and project them onto concentric spheres, producing geometrically grounded but per-view feature volumes. In Stage 2, we plug in our View-pair Correlation Transformer (VCT): for each sampled depth, we construct pairwise correlation volumes over all camera combinations and aggregate them into a sparsified, visibility-aware consistency volume via Top-$k$ attention. Stage 3 iteratively refines the depth map under the guidance of both the consistency volume and the context features. We design a lightweight context fuser to construct global semantic context. This design preserves RomniStereo's strong geometric modeling and recurrent refinement, while achieving explicit, reference-free multi-view reasoning and robust spherical coherence.

Extensive experiments on synthetic and real panoramas (OmniThings, OmniHouse, and Sunny) demonstrate that FreeOmniMVS consistently improves accuracy and seam coherence over spherical-sweeping, rectified, and hybrid baselines under identical calibrations, depth ranges, and evaluation masks. Qualitative results further show reduced cross-camera artifacts and better thin-structure preservation.

In summary, our major contributions are three-fold: (I) We introduce \textbf{FreeOmniMVS}, a reference-free omni-MVS framework that casts depth estimation as multi-view consistency maximization and yields visibility-aware, globally coherent depth estimation. (II) We propose the VCT, which explicitly models correlations across all camera pairs and uses attention with Top-$k$ sparsification to eliminate unreliable evidence while keeping aggregation scalable. (III) Experimental results on clean and occlusion-augmented benchmarks suggest that FreeOmniMVS maintains competitive accuracy while offering better tolerance to locally degraded views.

\section{Related Work}
\label{sec:relatedwork}

\subsection{Multi-Fisheye Stereo Matching}

\noindent\textbf{Spherical sweeping-based methods.} These approaches construct cost volumes by projecting multi-view features onto a series of concentric spherical depth planes, allowing the network to perform matching in the spherical domain rather than in distorted image space. Pioneered by SweepNet \cite{won2019sweepnet} and established by OmniMVS \cite{won2019omnimvs,won2020end}, this line of work explicitly incorporates the geometry of omnidirectional projection. Later variants such as IcoSweepNet \cite{komatsu2020360}, S-OmniMVS \cite{chen2023s}, and RomniStereo \cite{jiang2024romnistereo} introduced refined representations using icospheres or deformable spherical convolutions to improve geometric fidelity.

\noindent\textbf{Epipolarity-based reconstruction.} Another direction seeks to reduce distortions by transforming fisheye imagery into domains that are more compatible with conventional stereo matching. MODE \cite{li2022mode} and OmniVidar \cite{xie2023omnividar} adopt multi-perspective or Cassini projections, enabling rectified feature alignment across adjacent cameras. Such methods are computationally efficient and easier to train, but they process view pairs independently, thereby neglecting global multi-view relationships and assuming quasi-planar epipolar geometry even under extreme fisheye distortion.

\noindent\textbf{Hybrid approaches.} OmniStereo \cite{deng2025omnistereo} represents an intermediate paradigm that combines rectified stereo matching with partial spherical sweeping, achieving a better balance between efficiency and geometric accuracy. Nevertheless, such hybrid formulations still rely on learned feature fusion rather than explicit modeling of inter-view consistency, leaving cross-view coherence largely implicit.

In summary, these methods fail to explicitly exploit geometric relationships between multiple views, rendering them less capable of capturing global dependencies, visibility, or scale changes. By contrast, FreeOmniMVS aggregates pair-wise correlations into a robust, visibility-aware, and global consensus, making it tolerant to occlusions, partial overlaps, and varying baselines.

\begin{figure*}[t]
  \centering
  \includegraphics[width=0.98\textwidth]{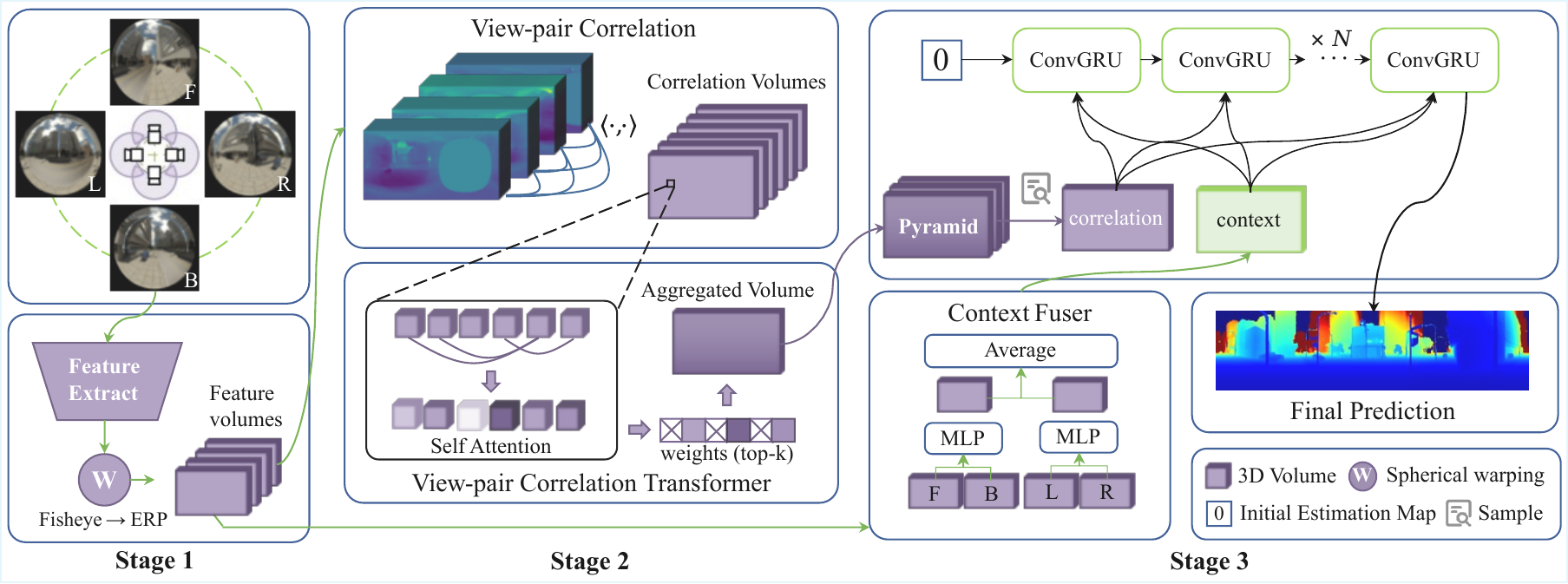}
  \vspace{-10pt}
  \caption{\textbf{Overall architecture of FreeOmniMVS.} Given four fisheye images, \textbf{Stage 1} extracts multi-scale unary features and performs OmniMVS-style spherical sweeping to build per-view feature volumes on the ERP sphere. \textbf{Stage 2} applies the proposed VCT to construct pairwise correlation volumes over all camera combinations and aggregates them, via sparse Top-$k$ attention, into a visibility-aware consistency volume $\mathcal{C}_{\text{fused}}$. \textbf{Stage 3} uses a lightweight context fuser to build a global context volume and feeds both context and consistency features into a RAFT-Stereo-style recurrent updater, which starts from a zero inverse-depth map and iteratively refines it, followed by convex upsampling to obtain the final high-resolution omnidirectional depth.}
  \label{fig:framework}
  \vspace{-10pt}
\end{figure*}

\subsection{Explicit Cost Volume Aggregation}

Early stereo and multi-view methods explicitly modeled cross-view similarity by constructing cost volumes that encode geometric correspondence. These approaches laid the foundation for modern depth estimation through convolutional regularization and hierarchical refinement. Methods such as PSMNet \cite{chang2018pyramid}, GA-Net \cite{zhang2019ga}, AANet \cite{xu2020aanet}, CFNet \cite{shen2021cfnet}, and PCW-Net \cite{shen2022pcw} explicitly model left-right feature similarity by constructing 4D cost volumes and regularizing them with 3D CNNs or guided aggregation modules. Subsequent iterative and hybrid designs, including RAFT-Stereo \cite{lipson2021raft}, CREStereo \cite{li2022practical}, MoCha-Stereo \cite{chen2024mocha}, Selective-Stereo \cite{wang2024selective}, IGEV \cite{xu2023iterative}, IGEV++ \cite{xu2025igev++}, MonSter++ \cite{Cheng_2025_CVPR}, and FoundationStereo \cite{wen2025stereo}, use recurrent or attention operators to achieve superior global consistency and computational efficiency.

Learning-based methods, such as MVSNet \cite{yao2018mvsnet}, R-MVSNet \cite{yao2019recurrent}, AA-RMVSNet \cite{wei2021aa}, D2HC-RMVSNet \cite{yan2020dense}, Vis-MVSNet \cite{zhang2020visibility}, UCS-Net \cite{cheng2020deep}, CasMVSNet \cite{gu2020cascade}, EPP-MVSNet \cite{ma2021epp}, and PatchmatchNet \cite{wang2021patchmatchnet}, extended these ideas to multiple views through differentiable homography warping and variance-based cost aggregation, explicitly encoding multi-view consistency along hypothesized depth planes. This geometric correspondence remains the foundation for most learning-based MVS pipelines.

\subsection{Transformer in Depth Estimation}

Recent works \cite{ding2022transmvsnet,shi2023raymvsnet++,wu2024gomvs,cao2022mvsformer} introduce transformer architectures to enhance global context modeling and cross-view interaction. TransMVSNet \cite{ding2022transmvsnet} integrates convolutional backbones with a feature-matching transformer to jointly model intra-view context and inter-view correspondence. Subsequent works extend this paradigm with transformers for long-range spatial reasoning and geometric consistency, such as MVSTR \cite{zhu2021multi} and MVSTER \cite{wang2022mvster}. In stereo matching, CroCo v2 \cite{weinzaepfel2023croco} adopts a pure transformer backbone to learn geometric priors via cross-view completion without explicit cost volumes, while FoundationStereo \cite{wen2025stereo} incorporates transformer attention into convolutional cost-volume aggregation. Although these approaches improve cross-view geometry modeling, they typically remain reference-centric or operate on limited view pairings. In contrast, we formulate a symmetric, reference-free consensus by applying attention over all view-pair correlation responses and introduce sparse Top-$k$ selection to discard unreliable pairs, enabling visibility-aware and scalable multi-view consistency for multi-fisheye rigs.

\section{Methodology}

\subsection{Preliminaries of Multi-fisheye Stereo Matching}

Multi-view stereo methods typically assume a \emph{reference view} as the geometric anchor. 
Given a reference image $I_r$ and a set of source views $\{I_s\}$, traditional MVS warps source features into the reference frustum via planar homography and computes pairwise feature similarity to form a cost volume, from which the optimal depth is obtained by maximizing confidence along the depth axis.

This reference-centric formulation becomes problematic for multi-fisheye rigs. 
Fisheye cameras exhibit strong radial distortion and each covers only a limited field of view, while large inter-view baselines lead to severe spatial discontinuities under planar homography warping.
To address this, omnidirectional MVS replaces planar homographies with spherical sweeping: depth hypotheses are represented by concentric spheres centered at the rig origin, and features from all cameras are re-sampled onto the same spherical surface:
\begin{equation}
\mathbf{p}_s \sim \Pi_{\text{fisheye}}\!\left(\mathbf{R}_s^{\top}(d\,\mathbf{n}_{\text{sph}}(\mathbf{x}) - \mathbf{t}_s)\right),
\end{equation}
where $\mathbf{n}_{\text{sph}}(\mathbf{x})$ denotes the unit ray direction of the spherical pixel.
This produces aligned feature maps $f_s(d,\mathbf{x})$ across cameras.
However, spherical sweeping still typically evaluates how well each source matches a chosen reference, which is suboptimal when no camera observes the entire $360^\circ$ scene.

\noindent\textbf{Our Idea.}
Let $\{f_v(d,\mathbf{x})\}_{v=1}^{N}$ denote the features of all fisheye cameras projected onto a common spherical surface at depth $d$.
Instead of reference-to-source matching, we measure the mutual consistency among all visible view pairs:
\begin{equation}
\label{eq:consistency}
\mathcal{C}(d,\mathbf{x}) = \sum_{i<j} \langle f_i(d,\mathbf{x}), f_j(d,\mathbf{x}) \rangle .
\end{equation}
Intuitively, the correct depth maximizes cross-view agreement by aligning observations of the same 3D point, and the optimal depth $d^*(\mathbf{x})$ is therefore obtained by maximizing $\mathcal{C}(d,\mathbf{x})$ over $d$.
This reformulation transforms omnidirectional MVS from reference-based photometric matching to global multi-view consistency maximization, removing the dependence on an arbitrary reference camera and forming the basis of our View Correlation Fusion module.

\subsection{The Proposed FreeOmniMVS Framework}
\label{sec:framework_sec}

\noindent\textbf{Overview.} An overview of FreeOmniMVS is illustrated in Fig.~\ref{fig:framework}. We build our model on the three-stage omnidirectional stereo backbone of RomniStereo \cite{jiang2024romnistereo}, and replace its adaptive feature volume with our reference-free multi-view consensus module. Given four fisheye images $\{I_v\}_{v=1}^{4}$, the first stage follows an OmniMVS-style spherical sweeping scheme \cite{won2020end}. Each view is first encoded by a shared 2D CNN into multi-scale unary features, which are then projected onto a set of concentric spheres aligned with the camera rig. This spherical sweeping yields geometrically grounded, per-view feature volumes on the equirectangular sphere, from which a shallow regression head produces an initial omnidirectional depth prior $D_{\text{coarse}}$.

In the second stage, we plug in our View-pair Correlation Transformer. For each sampled depth hypothesis and each equirectangular location, we gather the swept features from all cameras and construct pairwise correlation volumes over all view combinations. These per-pair correlation vectors are then fed into the transformer, which uses attention with a hard Top-$k$ sparsification to aggregate them into a visibility-aware consistency volume. In this way, unreliable pairs caused by occlusions or missing overlaps are down-weighted or discarded, and the resulting consistency cues no longer depend on any designated reference view.

The third stage performs recurrent refinement on the ERP grid, but operates entirely in the inverse-depth domain. We initialize the current inverse depth estimate $d_0$ as a zero map, and obtain the initial hidden state of a 2D convolutional GRU by applying a $1\times1$ convolution to the context slice corresponding to this initialization. At iteration $i$, given the current inverse depth $d_i$, we bilinearly sample a context feature map from the context volume and, in parallel, extract a correlation feature map by gathering responses from the consistency volume around $d_i$. For each pixel, we stack correlation entries at the current inverse depth and a fixed set of neighboring depth bins across all pyramid levels. The sampled context and correlation features are fed into the GRU to update its hidden state, from which a residual inverse-depth map $\Delta d_i$ and a convex upsampling mask are predicted. The estimate is then updated as $d_{i+1} = d_i + \Delta d_i$. The recurrent updates are computed at half resolution, and the final full-resolution depth is recovered from the last inverse-depth estimate and the learned convex upsampling mask.

To supply rich semantic cues to this recurrent stage, our lightweight context fuser first aggregates multi-scale 2D ERP features and fuses information across views into a unified context volume. This allows the GRU to refine the depth field under the joint guidance of both the visibility-aware consistency volume and the globally informed context features.

\subsection{View Correlation Construction}
\label{sec:vcc}

Let $\{f_v\in\mathbb{R}^{C\times D\times H\times W}\}_{v=1}^{N}$ denote the feature volumes extracted from the fisheye images after geometric alignment onto the $360^\circ$ domain with $N$ views. In our setting $N=4$, leading to $V_p = \binom{4}{2} = 6$ unordered view pairs. For each pair $(i,j)$, we construct a correlation volume that measures the similarity of the two feature volumes at every depth hypothesis $d$ and spatial location $\mathbf{x}$. To obtain a numerically stable and scale-invariant similarity measure, we employ a normalized inner product across the channel dimension:
\begin{equation}
\label{eq:pair_corr}
\mathcal{C}_{ij}(d,\mathbf{x}) = \frac{1}{C}\sum_{c=1}^{C}\frac{f_{i,c}(d,\mathbf{x})f_{j,c}(d,\mathbf{x})}{\|f_i(d,\mathbf{x})\|_2\,\|f_j(d,\mathbf{x})\|_2},
\end{equation}
where $f_{v,c}$ denotes the $c$-th channel of view $v$, and $\|\cdot\|_2$ is the $\ell_2$ norm over channels. We compute Eq.~\eqref{eq:pair_corr} for all view pairs $(i,j)$ and stack the resulting correlation volumes along a new dimension, forming a multi-view correlation tensor $\mathbf{C} \in \mathbb{R}^{B \times V_p \times D \times H \times W}$, where $B$ is the batch size and $V_p$ is the number of view pairs. At each depth $d$ and location $\mathbf{x}$, the vector $\mathbf{c}(d,\mathbf{x}) \in \mathbb{R}^{V_p}$ collects the pairwise correlations from all view pairs and serves as the basic representation on which our transformer operates.

\subsection{View-pair Correlation Transformer}
\label{sec:VCT}

\begin{figure}[t]
  \centering
  \includegraphics[width=\linewidth]{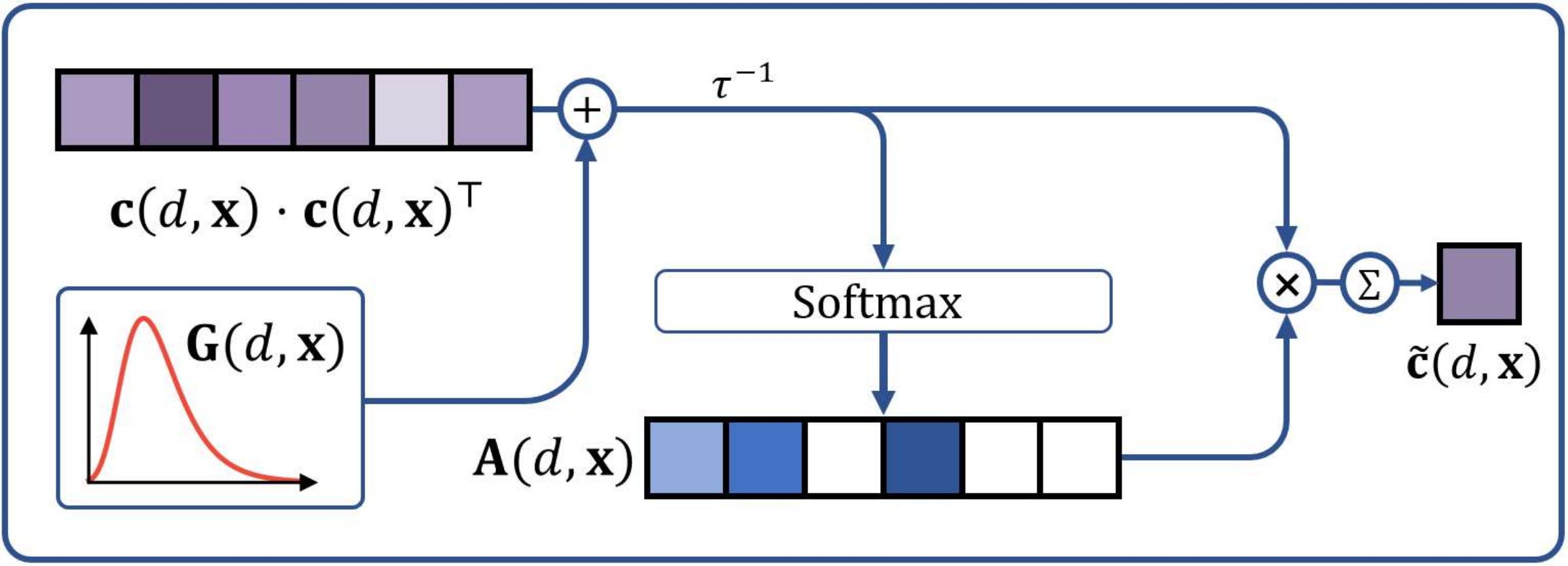}
  \vspace{-8pt}
  \caption{\textbf{Overview of the proposed VCT.} For each voxel $(d,\mathbf{x})$, VCT takes the vector of pairwise correlations across all camera pairs, builds a self-similarity matrix over view pairs, and applies attention with Top-$k$ sparsification, with Gumbel perturbation during training, to aggregate them into a fused scalar consistency score.}
  \label{fig:vct}
  \vspace{-12pt}
\end{figure}

Given the multi-view correlation tensor $\mathbf{C}$, our goal is to aggregate the $V_p$ pairwise responses into a single, robust consistency score for each depth hypothesis, as depicted in Fig.~\ref{fig:vct}. Intuitively, if a depth $d$ is correct at position $\mathbf{x}$, then the corresponding correlations from all visible view pairs should be mutually consistent; conversely, view pairs involving occluded or out-of-FoV observations should be down-weighted or discarded.

For a fixed $(d,\mathbf{x})$, we denote the correlation vector as $\mathbf{c} = \mathbf{c}(d,\mathbf{x}) \in \mathbb{R}^{V_p}$. We construct a self-similarity matrix that captures how strongly each view pair agrees with every other, and apply temperature-scaled softmax to obtain attention weights over view pairs:
\begin{equation}
\label{eq:attn_weights}
\begin{aligned}
\mathbf{S}(d,\mathbf{x}) &= \frac{1}{\sqrt{V_p}}\,\mathbf{c}(d,\mathbf{x})\,\mathbf{c}(d,\mathbf{x})^{\top}, \\
\mathbf{A}(d,\mathbf{x}) &= \operatorname{softmax}\!\left(\frac{\mathbf{S}(d,\mathbf{x}) + \mathbf{G}(d,\mathbf{x})}{\tau}\right),
\end{aligned}
\end{equation}
where $\tau > 0$ is a temperature parameter controlling the sharpness of the distribution, and $\mathbf{G}$ denotes an optional perturbation term used during training. The fused correlation vector is then obtained by $\tilde{\mathbf{c}}(d,\mathbf{x}) = \mathbf{A}(d,\mathbf{x})\,\mathbf{c}(d,\mathbf{x})$, and the final scalar consistency score $\mathcal{C}_{\text{fused}}(d,\mathbf{x})$ is computed by averaging the entries of $\tilde{\mathbf{c}}(d,\mathbf{x})$. Applying this at every $(d,\mathbf{x})$ yields a fused consistency volume $\mathcal{C}_{\text{fused}} \in \mathbb{R}^{B\times D\times H\times W}$, which is then used by the recurrent updater in the third stage to refine the final depth.

\noindent\textbf{Sparse Top-$k$ selection.} Not all view pairs are equally informative. Pairs involving occlusions, out-of-FoV regions, or heavily corrupted observations may introduce inconsistent correlations that should be ignored rather than merely down-weighted. To this end, we enforce sparsity in the attention weights by selecting only the Top-$k$ entries in each row of $\mathbf{S}(d,\mathbf{x})$. During training, we adopt a differentiable approximation based on Gumbel perturbations. Specifically, the matrix $\mathbf{G}(d,\mathbf{x})$ in Eq.~\eqref{eq:attn_weights} is sampled from a Gumbel distribution, and the subsequent softmax yields a stochastic but smooth approximation to a Top-$k$ operator over view pairs. This encourages the network to learn which view pairs are most reliable while preserving gradient flow.

\begin{table*}[t]
\caption{\textbf{Quantitative comparison with existing omnidirectional depth estimation methods on OmniThings, OmniHouse, and Sunny.} We report the percentage of pixels whose inverse-depth index error exceeds 1, 3, and 5 bins ($>1$, $>3$, $>5$), as well as MAE and RMSE. All models are either trained on OmniThings only (middle block) or additionally fine-tuned on OmniHouse and Sunny (bottom block). The best results are shown in \textbf{bold}, and the second best are \underline{underlined}.}
\vspace{-10pt}
\resizebox{\textwidth}{!}{%
\begin{tabular}{l|rrrrr|rrrrr|rrrrr}
\bottomrule
\multicolumn{1}{l|}{Dataset} & \multicolumn{5}{c|}{OmniThings} & \multicolumn{5}{c|}{OmniHouse} & \multicolumn{5}{c}{Sunny} \\
\cline{2-16}
\multicolumn{1}{l|}{Metric} & \multicolumn{1}{c}{$>1$} & \multicolumn{1}{c}{$>3$} & \multicolumn{1}{c}{$>5$} & \multicolumn{1}{c}{MAE} & \multicolumn{1}{c|}{RMSE} & \multicolumn{1}{c}{$>1$} & \multicolumn{1}{c}{$>3$} & \multicolumn{1}{c}{$>5$} & \multicolumn{1}{c}{MAE} & \multicolumn{1}{c|}{RMSE} & \multicolumn{1}{c}{$>1$} & \multicolumn{1}{c}{$>3$} & \multicolumn{1}{c}{$>5$} & \multicolumn{1}{c}{MAE} & \multicolumn{1}{c}{RMSE} \\
\toprule
\multicolumn{16}{l}{\it{Trained on OmniThings only}} \\
\midrule
OmniMVS~\cite{won2019omnimvs} & 47.72 & 15.12 & 8.91 & 2.40 & 5.27 & 30.53 & 10.29 & 6.27 & 1.72 & 4.05 & 27.16 & 6.13 & 3.98 & 1.24 & 3.09 \\
S-OmniMVS~\cite{chen2023s} & 28.03 & 10.40 & 6.33 & 1.48 & \textbf{3.68} & 18.86 & 8.05 & 4.90 & 1.06 & 2.41 & 17.19 & 6.03 & 3.89 & 1.11 & 3.60 \\
OmniMVS${}^+_{32}$~\cite{won2020end} & 20.70 & \textbf{8.18} & \underline{5.49} & \underline{1.37} & 4.11 & 19.89 & 5.89 & 3.99 & 1.30 & 2.64 & 13.57 & \textbf{4.81} & \textbf{3.10} & 0.88 & \textbf{2.56} \\
RomniStereo${}_{32}$~\cite{jiang2024romnistereo} & \underline{20.42} & 8.49 & 5.81 & 1.39 & 4.22 & \textbf{12.13} & \underline{4.73} & \underline{3.02} & \underline{0.80} & \textbf{1.85} & \textbf{12.28} & 5.59 & 3.79 & \textbf{0.80} & 2.68 \\
\textbf{FreeOmniMVS} & \textbf{19.62} & \underline{8.22} & \textbf{5.41} & \textbf{1.32} & \underline{4.10} & \underline{12.31} & \textbf{4.63} & \textbf{2.90} & \textbf{0.79} & \textbf{1.85} & \underline{12.37} & \underline{5.51} & \underline{3.75} & \textbf{0.80} & \underline{2.67} \\
\toprule
\multicolumn{16}{l}{\it{Finetuned on OmniHouse and Sunny}} \\
\midrule
OmniMVS-ft~\cite{won2019omnimvs} & 50.28 & 22.78 & 15.60 & 3.52 & 7.44 & 21.09 & 4.63 & 2.58 & 1.04 & 1.97 & 13.93 & 2.87 & 1.71 & 0.79 & 2.12 \\
S-OmniMVS-ft~\cite{chen2023s} & - & - & - & - & - & 6.99 & \underline{1.79} & \textbf{0.97} & \underline{0.42} & \textbf{1.06} & 6.66 & 2.18 & 1.40 & 0.47 & 1.98 \\
OmniMVS${}^+_{32}$-ft~\cite{won2020end} & 44.79 & 27.17 & 20.41 & 4.23 & 8.42 & 9.70 & 3.51 & 2.13 & 0.64 & 1.69 & 7.48 & 3.57 & 2.42 & 0.57 & 2.42 \\
RomniStereo${}_{32}$-ft~\cite{jiang2024romnistereo} & \underline{34.32} & \underline{19.76} & \underline{14.22} & \underline{2.81} & \underline{6.47} & \underline{6.02} & 2.49 & 1.73 & 0.49 & 1.31 & \textbf{5.19} & \textbf{1.98} & \textbf{1.23} & \textbf{0.36} & \textbf{1.55} \\
\textbf{FreeOmniMVS-ft} & \textbf{29.13} & \textbf{15.00} & \textbf{10.38} & \textbf{2.24} & \textbf{5.63} & \textbf{4.99} & \textbf{1.62} & \underline{0.99} & \textbf{0.36} & \textbf{1.06} & \underline{5.54} & \underline{2.01} & \underline{1.25} & \underline{0.37} & \underline{1.57} \\
\toprule
\end{tabular}}
\vspace{-10pt}
\label{tab:quan}
\end{table*}

\newcommand{\turnheightnew}{0.215\columnwidth}

\begin{figure*}[t]
\begin{tabular}{@{\hskip 1mm}c@{\hskip 1mm}c@{\hskip 1mm}c@{\hskip 1mm}c@{\hskip 1mm}c@{\hskip 1mm}c@{}}
{\rotatebox{90}{\hspace{1.0mm}\scriptsize OmniThings}} &
\includegraphics[height=\turnheightnew]{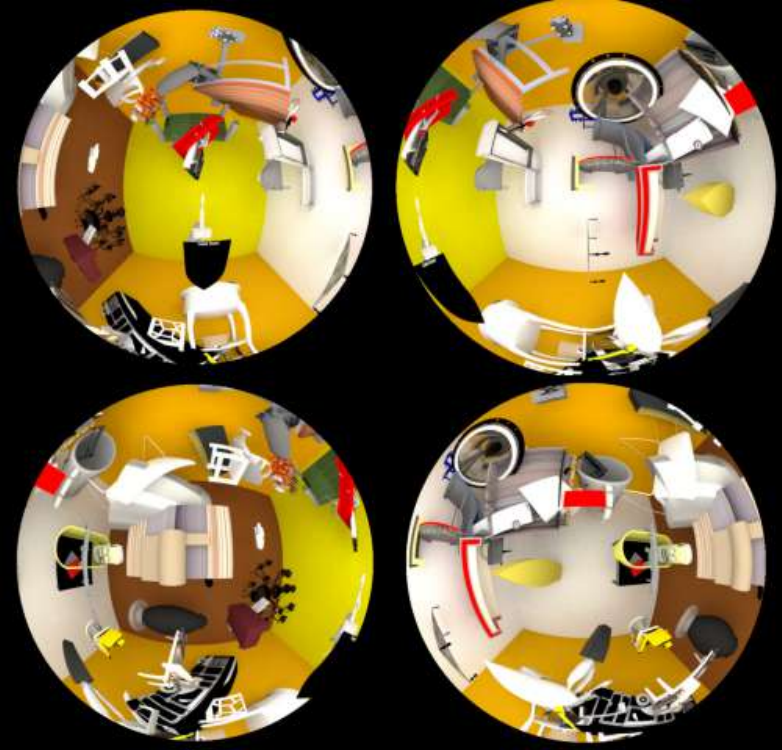} &
\includegraphics[height=\turnheightnew]{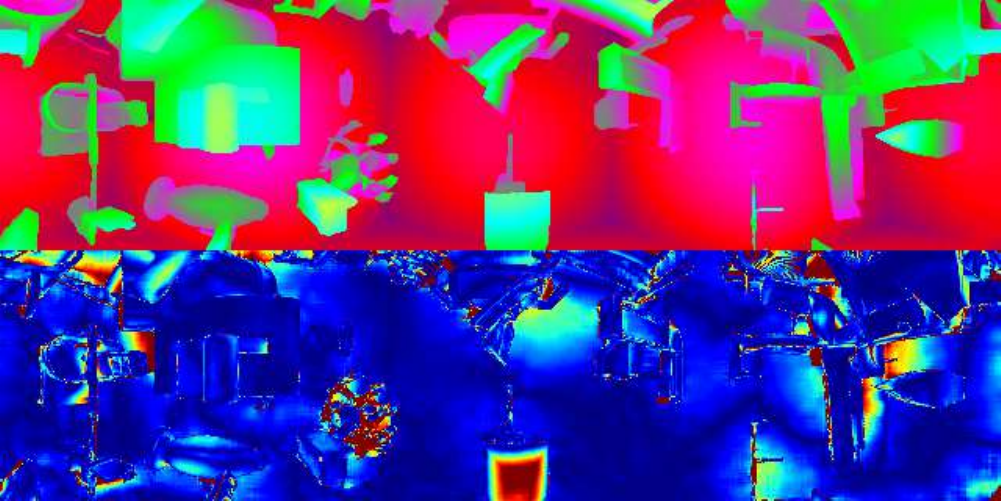} &
\includegraphics[height=\turnheightnew]{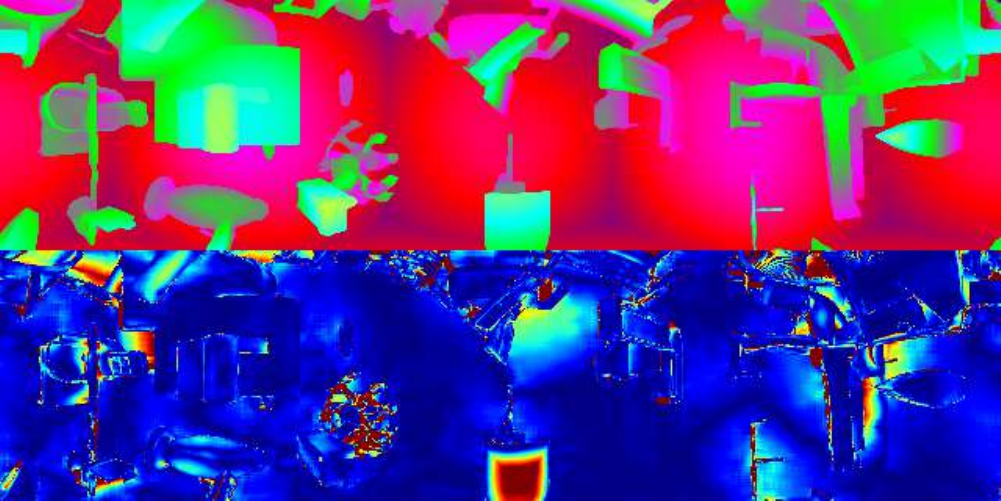} &
\includegraphics[height=\turnheightnew]{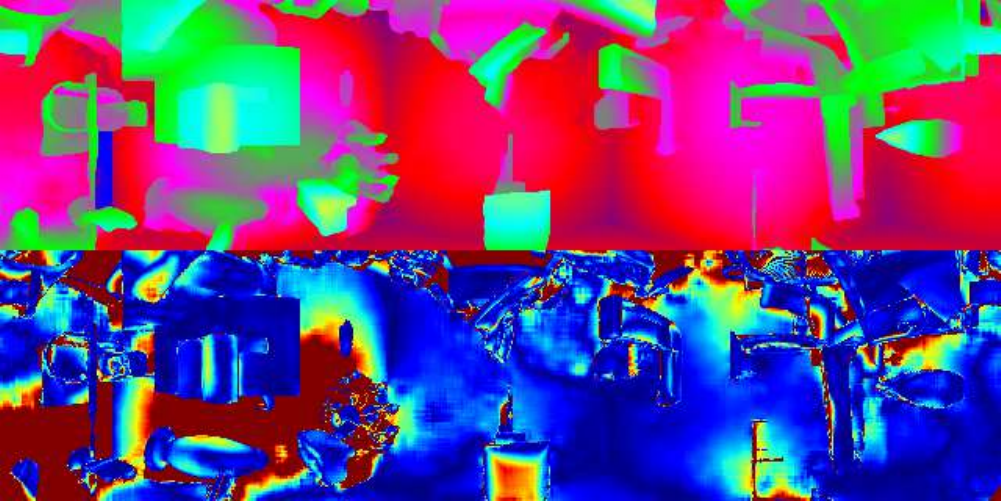} &
\includegraphics[height=\turnheightnew]{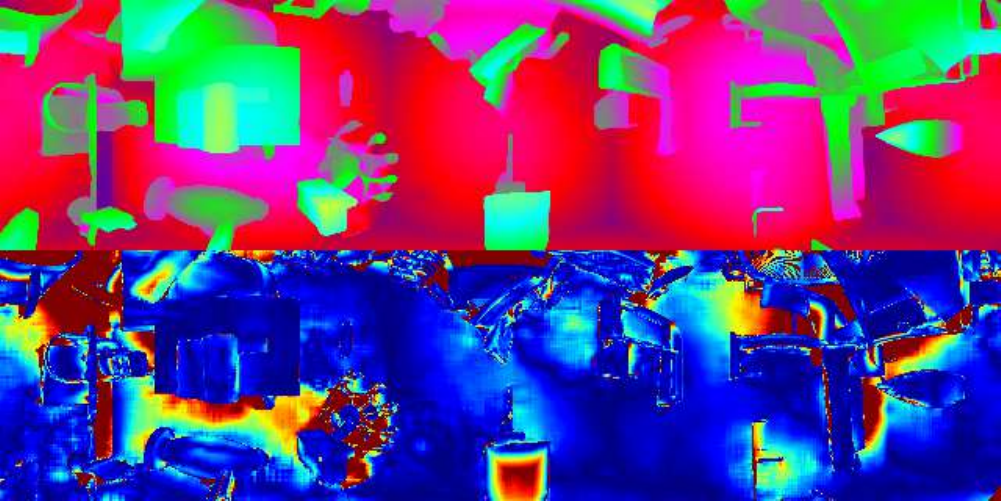} \\
{\rotatebox{90}{\hspace{1.6mm}\scriptsize OmniHouse}} &
\includegraphics[height=\turnheightnew]{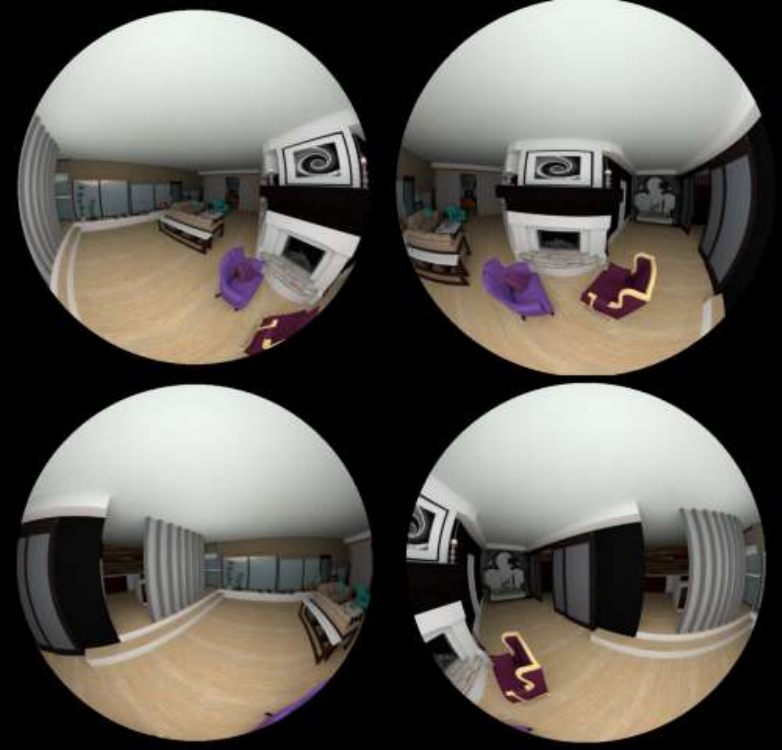} &
\includegraphics[height=\turnheightnew]{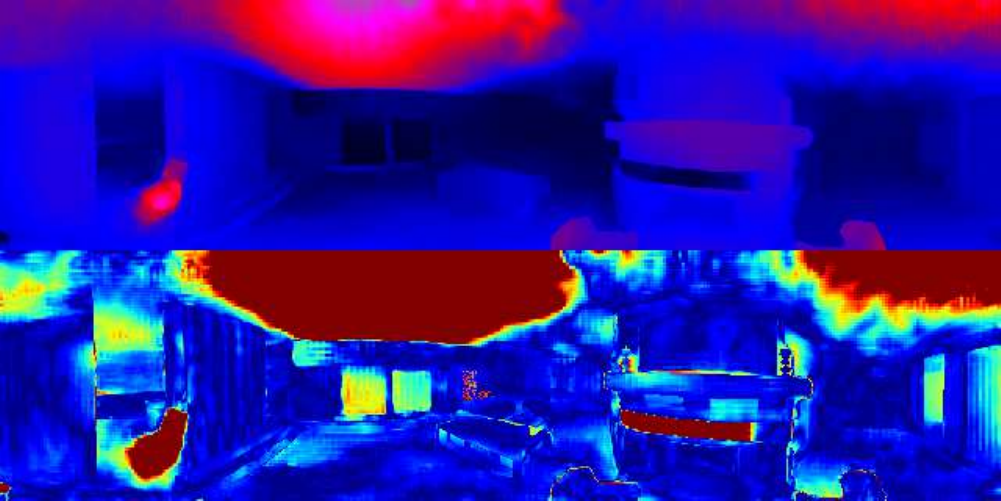} &
\includegraphics[height=\turnheightnew]{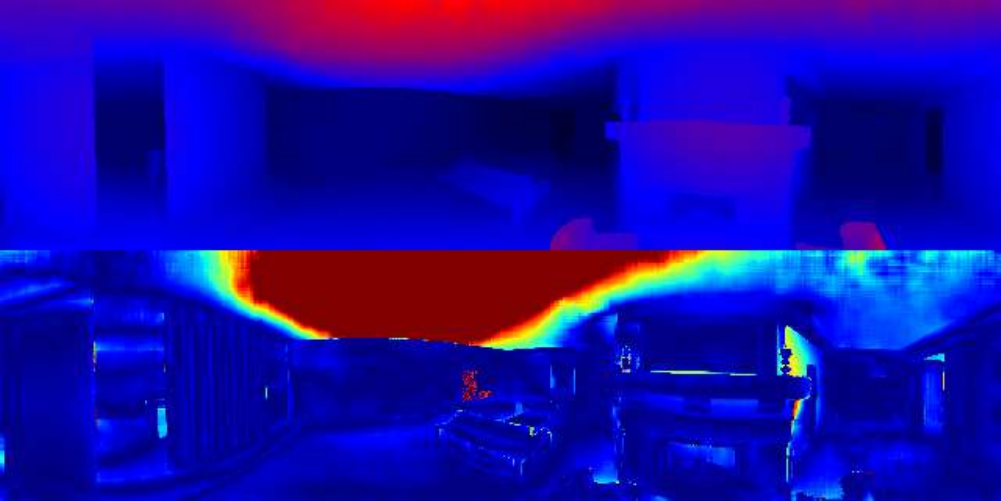} &
\includegraphics[height=\turnheightnew]{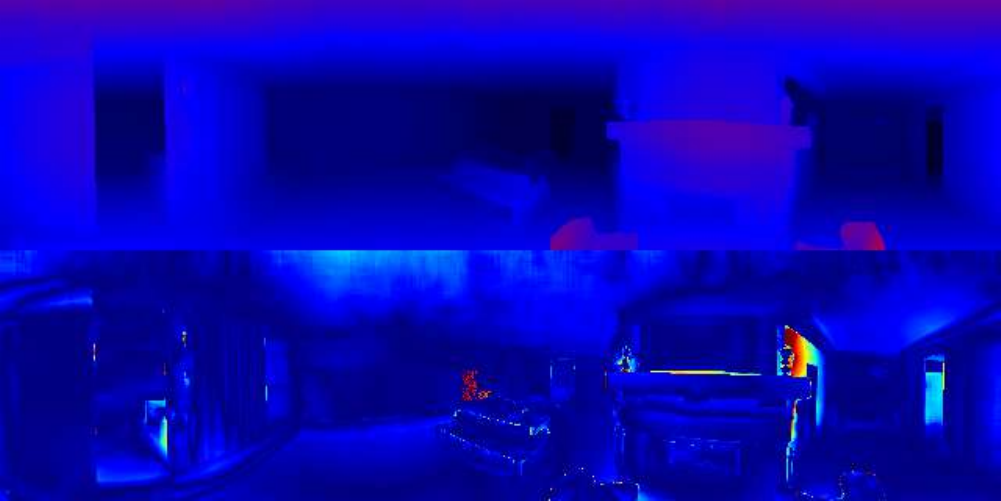} &
\includegraphics[height=\turnheightnew]{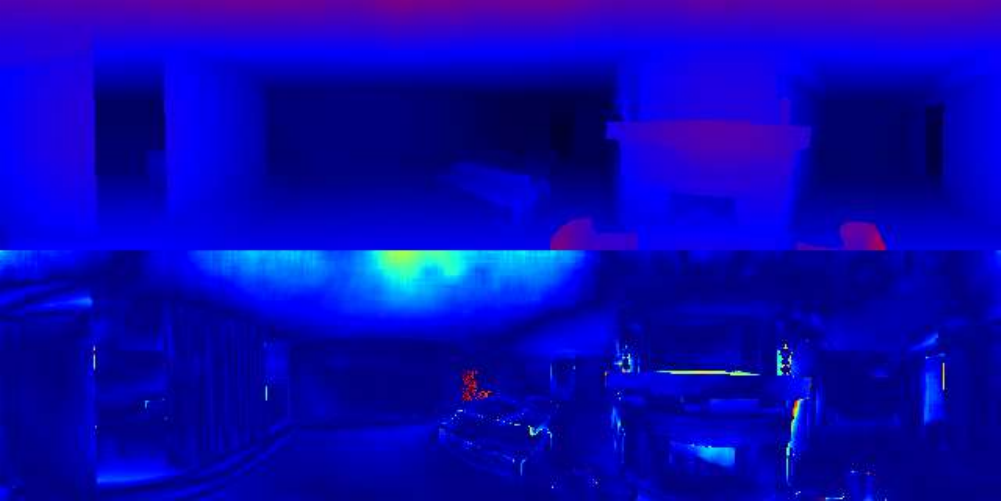} \\
{\rotatebox{90}{\hspace{5.0mm}\scriptsize Sunny}} &
\includegraphics[height=\turnheightnew]{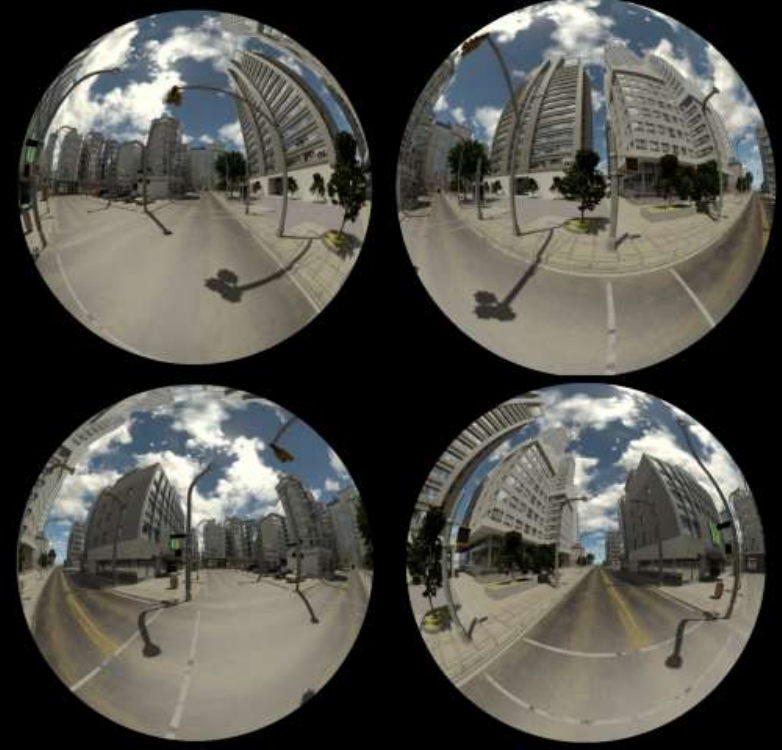} &
\includegraphics[height=\turnheightnew]{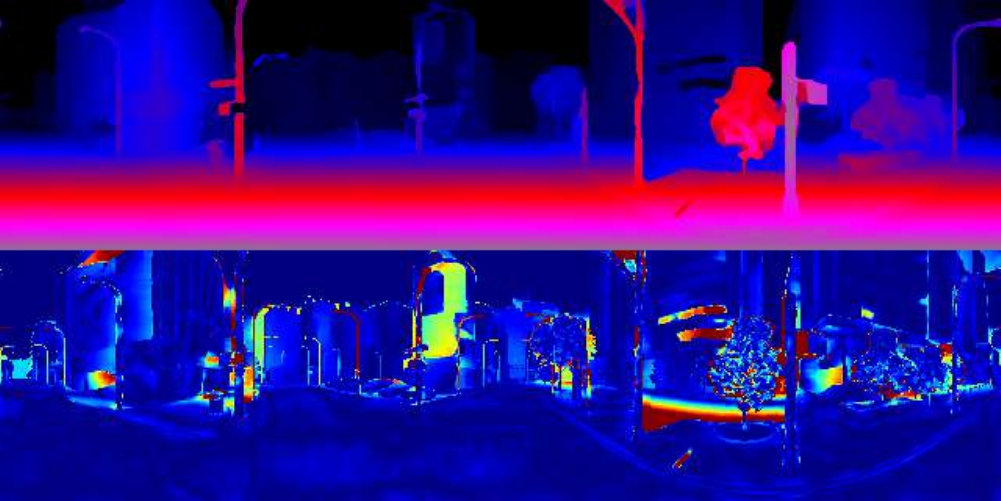} &
\includegraphics[height=\turnheightnew]{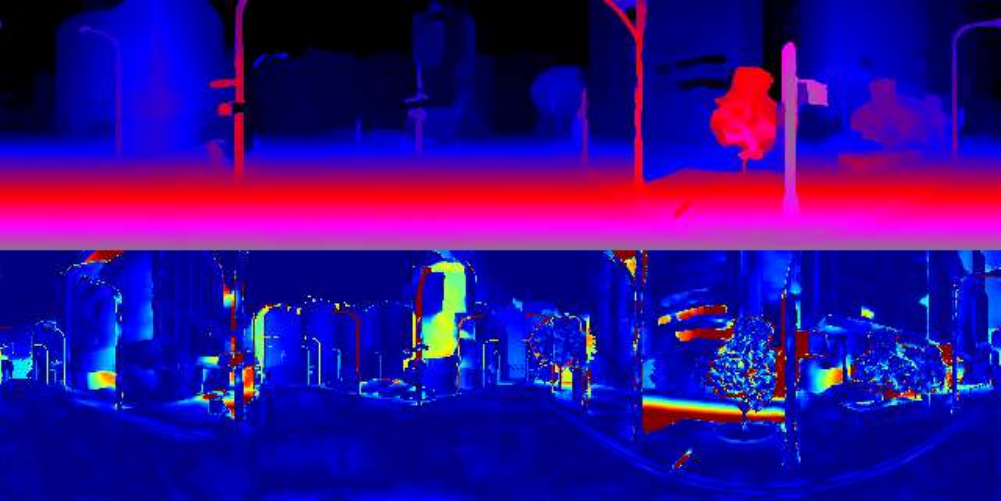} &
\includegraphics[height=\turnheightnew]{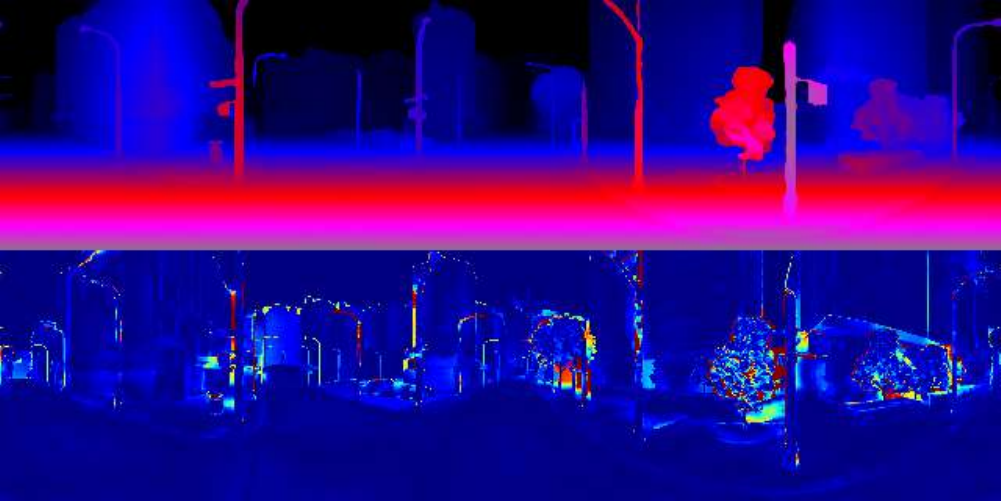} &
\includegraphics[height=\turnheightnew]{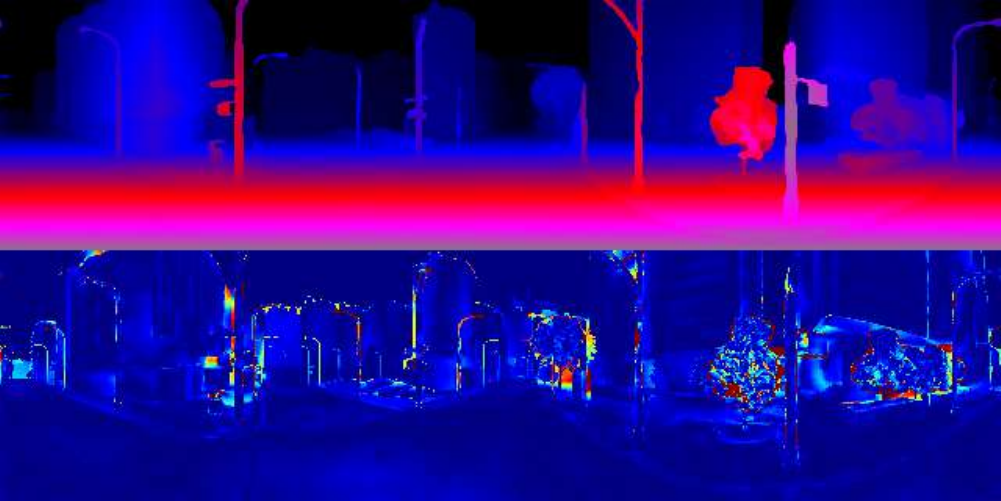} \\
& \scriptsize Input & \scriptsize RomniStereo${}_{32}$ & \scriptsize FreeOmniMVS & \scriptsize RomniStereo${}_{32}$-ft & \scriptsize FreeOmniMVS-ft \\
\end{tabular}
\caption{Qualitative comparison on clean test images from OmniThings, OmniHouse, and Sunny.}
\vspace{-10pt}
\label{fig:qualitative}
\end{figure*}

At inference time, we set $\mathbf{G}=0$ and explicitly apply a hard Top-$k$ mask on $\mathbf{S}(d,\mathbf{x})$ before the softmax, zeroing out all but the $k$ largest entries ($k=3$ in practice) in each row. In this way, our transformer aggregates only the most consistent view pairs and fully discards unreliable correlations, effectively emulating a visibility-aware mask without explicit geometric reasoning. The entire fusion process is parameter-free and operates locally at each voxel $(d,\mathbf{x})$, making it both efficient and easily integrable into existing omnidirectional depth estimation pipelines.

\subsection{Loss Function and Optimization}
\label{sec:loss}

We follow RomniStereo \cite{jiang2024romnistereo} and supervise all recurrent predictions of the inverse-depth map. Let $\mathbf{d}^{(i)} \in \mathbb{R}^{B\times H_{\text{ERP}}\times W_{\text{ERP}}}$ denote the inverse-depth estimate at the $i$-th iteration of the recurrent updater, and let $\mathbf{d}_{\text{gt}}$ be the corresponding ground-truth inverse-depth index map. Given $M$ iterations in total, we define the training loss as an exponentially weighted $\ell_1$ regression over all intermediate predictions:
\begin{equation}
\mathcal{L} = \sum_{i=1}^{M} \gamma^{M-i}\frac{1}{|\Omega|}\sum_{\mathbf{x}\in\Omega}\left\lVert \mathbf{d}_{\text{gt}}(\mathbf{x}) - \mathbf{d}^{(i)}(\mathbf{x}) \right\rVert_1,
\end{equation}
where $\Omega$ denotes the set of valid ERP pixels and $\gamma \in (0,1)$ controls the relative importance of early versus late iterations. We set $\gamma=0.9$ in all experiments.

\section{Experiments}
\label{sec:exp}

\subsection{Experimental Setup}
\label{sec:exp_setup}

\noindent\textbf{Datasets.} We evaluate FreeOmniMVS on three synthetic omnidirectional benchmarks widely used in prior work: OmniThings, OmniHouse, and Sunny. OmniThings contains generic objects under diverse lighting and geometric configurations, OmniHouse focuses on indoor environments, and Sunny covers driving scenes under clear weather. For all experiments we follow the official train, validation, and test splits from previous works and report results on the test sets of all three datasets.

To further assess robustness under view degradation, we construct corrupted test sets on OmniThings, OmniHouse, and Sunny. For each test sample and each fisheye camera, we use a deterministic random state derived from the sample and camera indices; with probability $0.3$ we inject between one and four circular occlusions into the image. Each occlusion has a radius sampled uniformly in $[0.01, 0.1]$ of the minimum image dimension and a randomly chosen type: noise or blur. Noise occlusions add random intensity perturbations within the circular mask, while blur occlusions replace the masked region with a Gaussian-blurred version of the image (kernel size $15\times15$, standard deviation $5$). The corrupted images share the same depth ground truth as the clean ones and allow us to evaluate how well the model maintains multi-view consistency when some views are locally out of focus or heavily contaminated.

\noindent\textbf{Training details.} Unless otherwise stated, all models are trained in PyTorch on a single NVIDIA RTX 3090 GPU with a batch size of 2. We adopt the same optimization setup as RomniStereo \cite{jiang2024romnistereo}: AdamW as the optimizer and a OneCycle learning-rate scheduler with a maximum learning rate of $5\times10^{-4}$. Following the common protocol, we first pre-train the network on OmniThings for 30 epochs and then fine-tune it for 15 epochs on the union of OmniHouse and Sunny. All stages of the backbone and our proposed modules are optimized jointly using the loss in Sec.~\ref{sec:loss}.

\noindent\textbf{Metrics.} We evaluate depth prediction quality in the inverse-depth index domain, using exactly the same metrics as RomniStereo for fair comparison. Depth is discretized into $N$ predefined inverse-depth indices, and we report the percentage of pixels whose index error exceeds 1, 3, and 5 bins, together with the mean absolute error and root mean squared error. All metrics are computed over valid ERP pixels using the same evaluation masks and index discretization as in prior work.

\subsection{Comparative Results}
\label{sec:comparison}

\noindent\textbf{Accuracy on clean benchmarks.} We first compare FreeOmniMVS with representative omnidirectional depth estimators, including the spherical-sweeping methods OmniMVS \cite{won2019omnimvs}, S-OmniMVS \cite{chen2023s}, OmniMVS${}^+_{32}$ \cite{won2020end}, and the recent RomniStereo${}_{32}$ \cite{jiang2024romnistereo}. For a fair comparison, all competitors are trained on OmniThings only or further fine-tuned on OmniHouse and Sunny following the same protocol as Sec.~\ref{sec:exp_setup}. Quantitative results are summarized in Tab.~\ref{tab:quan}, and qualitative comparisons are shown in Fig.~\ref{fig:qualitative}.

When trained on OmniThings only, FreeOmniMVS achieves the best or second-best score on almost all metrics. Compared to the RomniStereo${}_{32}$ backbone, our reference-free multi-view reasoning yields consistently lower errors: on OmniThings, we reduce MAE from $1.39$ to $1.32$ and the $>5$ index error from $5.81\%$ to $5.41\%$; on OmniHouse, we improve $>3$ from $4.73\%$ to $4.63\%$ and $>5$ from $3.02\%$ to $2.90\%$, while slightly lowering RMSE from $4.22$ to $4.10$; on Sunny, we maintain the strong MAE of RomniStereo ($0.80$) and slightly improve RMSE from $2.68$ to $2.67$.

After fine-tuning on OmniHouse and Sunny, the gains become more pronounced. On OmniThings, FreeOmniMVS-ft reduces the $>5$ index error from $14.22\%$ to $10.38\%$ and MAE from $2.81$ to $2.24$ compared to RomniStereo${}_{32}$-ft, corresponding to about a 20\% reduction in MAE. On OmniHouse, we further cut $>3$ from $2.49\%$ to $1.62\%$ and MAE from $0.49$ to $0.36$ (about 26\% relative improvement), while lowering RMSE from $1.31$ to $1.06$. On Sunny, FreeOmniMVS-ft closely matches RomniStereo${}_{32}$-ft, staying within $0.03$ on all metrics. Fig.~\ref{fig:qualitative} shows example samples with the predicted inverse-depth index maps and error maps of different models on the three benchmarks.



\begin{figure}[t]
  \centering
  \includegraphics[width=0.7\linewidth]{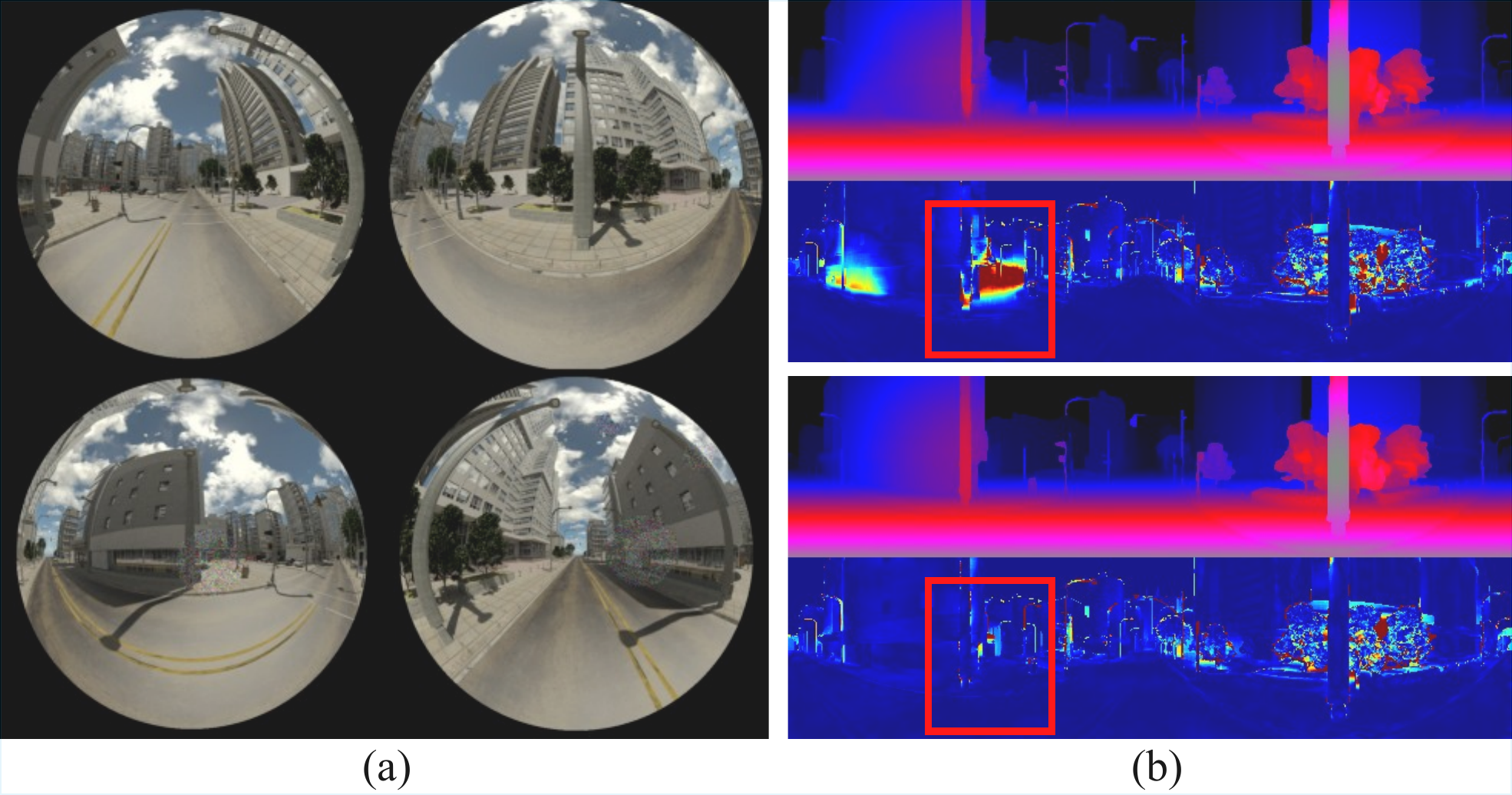}
  \includegraphics[width=0.7\linewidth]{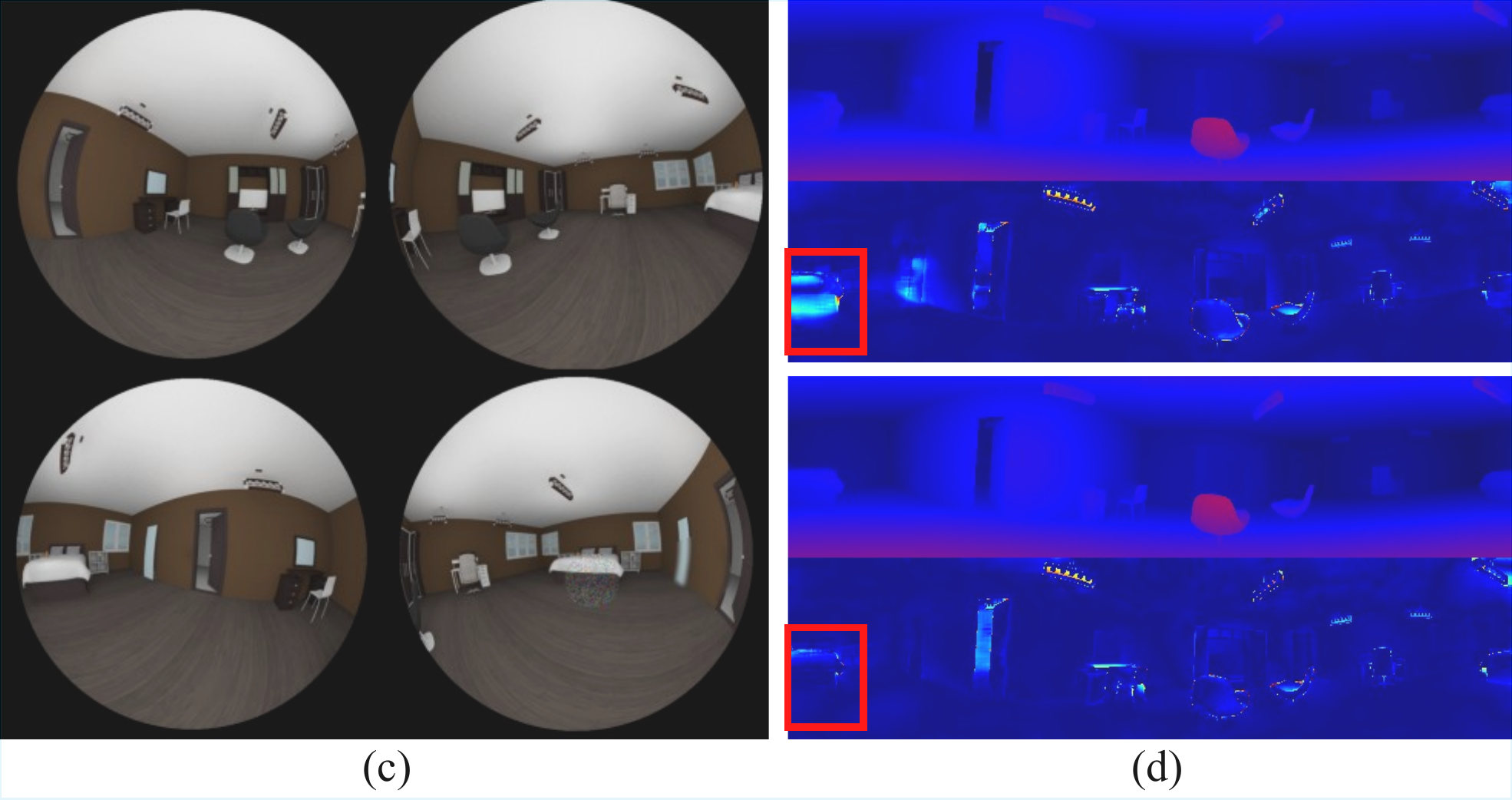}
  \vspace{-10pt}
  \caption{Corner cases on occlusion-augmented OmniHouse and Sunny. (a,c) Fisheye inputs with strong local blur or noise. (b,d) Depth predictions from RomniStereo${}_{32}$-ft (top) and FreeOmniMVS-ft (bottom). Our method better preserves thin structures and reduces large local errors around degraded regions.}
  \label{fig:corner_cases}
  \vspace{-13pt}
\end{figure}

\noindent\textbf{Robustness on degraded views.} To evaluate robustness under partial view degradation, we test on the occlusion-augmented versions of OmniHouse and Sunny. Here, each fisheye view may contain one to four randomly placed circular regions corrupted by strong blur or noise. As such local failures can lead to severe global artifacts, we focus on the RMSE metric. As shown in Tab.~\ref{tab:occ_rmse}, FreeOmniMVS-ft substantially outperforms RomniStereo${}_{32}$-ft: on OmniHouse-Occ, RMSE drops from $2.90$ to $1.71$ (about 41\% reduction), and on Sunny-Occ it decreases from $2.01$ to $1.61$ (around 20\% reduction). Qualitative corner cases in Fig.~\ref{fig:corner_cases} show that our reference-free VCT can effectively ignore corrupted view pairs and preserve thin structures and cross-camera consistency, whereas the baseline often exhibits large local errors around blurred or noisy regions.

\begin{table}[t]
\centering
\small
\setlength{\tabcolsep}{15pt}
\caption{RMSE on occlusion-augmented test sets (lower is better). FreeOmniMVS-ft significantly reduces the error compared to RomniStereo${}_{32}$-ft, especially on OmniHouse-Occ, where RMSE drops from 2.90 to 1.71.}
\vspace{-10pt}
\begin{tabular}{l|cc}
\bottomrule
Dataset (Occ) & OmniHouse & Sunny \\
\midrule
RomniStereo${}_{32}$-ft~\cite{jiang2024romnistereo} & 2.90 & 2.01 \\
\textbf{FreeOmniMVS-ft} & 1.71 & 1.61 \\
\toprule
\end{tabular}
\vspace{-8pt}
\label{tab:occ_rmse}
\end{table}

\subsection{Ablation Study}
\label{sec:ablation}

We conduct ablation experiments on OmniHouse, Sunny, and their occlusion-augmented versions to analyze the impact of the proposed VCT, Top-$k$ sparsification, and the context fuser. Results are summarized in Tab.~\ref{tab:ablation}.

\noindent\textbf{Effect of Top-$k$ sparsification.} We first vary the Top-$k$ sparsification strategy while keeping VCT and the context fuser enabled. Using very small $k$ leads to noticeable degradation: with $k=1$, MAE on OmniHouse rises to $1.04$ and $>1$ to $13.14\%$, and Sunny similarly suffers. Setting $k=2$ alleviates this issue but still lags behind the stronger configurations. Removing Top-$k$ entirely and aggregating all view pairs improves both clean and occluded performance compared to $k=1$ or $k=2$, suggesting that overly aggressive sparsification can discard useful evidence. Our full model, with a moderate Top-$k$ and Gumbel-based training, achieves the best results on both clean and corrupted data.

\noindent\textbf{Effect of the View-pair Correlation Transformer.} We remove the transformer and feed only the per-view context to the recurrent updater. On clean OmniHouse and Sunny, the MAE remains close to the full model. However, robustness degrades drastically once views are corrupted: on OmniHouse-Occ, RMSE jumps from $1.71$ to $2.85$, and on Sunny-Occ from $1.61$ to $3.98$. This indicates that while the backbone can still learn reasonable depth on clean images without explicit pairwise reasoning, VCT is crucial for reliably handling localized failures and occlusions by filtering out inconsistent view pairs.

\noindent\textbf{Effect of the context fuser.} Finally, we examine the effect of removing the context fuser, that is, using only VCT-based consistency without globally fused semantic context. On clean OmniHouse and Sunny, the absence of context fusion leads to a clear drop in accuracy. On the occlusion-augmented sets, RMSE on OmniHouse-Occ rises from $1.71$ to $2.04$, and on Sunny-Occ from $1.61$ to $2.83$. These results show that the context fuser and VCT play complementary roles: VCT provides visibility-aware pair selection, while the context fuser supplies globally consistent semantic cues.

\begin{table}[t]
\centering
\caption{Ablation study on OmniHouse, Sunny, and their occlusion-augmented versions.}
\vspace{-10pt}
\resizebox{\linewidth}{!}{%
\footnotesize
\begin{tabular}{l|cc|cc|c|c}
\bottomrule
Dataset & \multicolumn{2}{c|}{OH} & \multicolumn{2}{c|}{Sn} & OH (Occ) & Sn (Occ) \\
\midrule
Metric & $>1$ & MAE & $>1$ & MAE & RMSE & RMSE \\
\hline
top-k ($k=1$) & 13.14 & 1.04 & 16.19 & 1.11 & 2.13 & 3.04 \\
top-k ($k=2$) & 13.47 & 0.97 & 14.21 & 1.02 & 2.01 & 2.80 \\
w/o top-k & 12.74 & 0.94 & 13.52 & 0.89 & 1.95 & 2.74 \\
w/o VCT & 12.17 & 0.79 & 12.33 & 0.80 & 2.85 & 3.98 \\
w/o context fuser & 12.39 & 0.93 & 12.98 & 0.84 & 2.04 & 2.83 \\
full model & 12.31 & 0.79 & 12.37 & 0.80 & 1.71 & 1.61 \\
\toprule
\end{tabular}}
\vspace{-8pt}
\label{tab:ablation}
\end{table}




\section{Conclusion and Future Work}
\label{sec:conclusion}

In this paper, we presented \textbf{FreeOmniMVS}, a reference-free framework that casts omnidirectional depth regression as multi-view consistency maximization rather than relying on a reference view. With RomniStereo as backbone, we proposed a View-pair Correlation Transformer and a lightweight context fuser, modeling pairwise correlations across all fisheye cameras and fused global context without requiring any single view. Experiments on synthetic benchmarks and occlusion-augmented test sets showed that this design achieves competitive or better accuracy without increasing the model size. Our approach significantly improves robustness under local blur and noise, while generalizing well to real-world fisheye data.

\noindent\textbf{Future Work.} Future work may extend this paradigm to larger camera arrays and dynamic scenes, and explore joint learning of geometry and downstream perception tasks under the same reference-free formulation.

\bibliographystyle{IEEEtran}
\bibliography{main}

\end{document}